\documentclass[11pt,a4paper]{article}
\usepackage[hyperref]{emnlp-ijcnlp-2019}
\usepackage{times}
\usepackage{latexsym}
\usepackage{amsmath}
\usepackage{amssymb}
\usepackage{amsthm}
\usepackage{url}
\usepackage{qtree}
\usepackage{tabularx}
\usepackage{graphicx}
\usepackage{booktabs}
\newtheorem{prop}{Proposition}

\aclfinalcopy 

\newcommand{\ignore}[1]{}

\newcommand{\coo}{\overline{\textrm{COO}}}
\newcommand{\pname}{$\coo\ $}

\usepackage{xargs} 
\usepackage[colorinlistoftodos,prependcaption,textsize=tiny]{todonotes}

\newcommand{\reducedstrut}{\vrule width 0pt height .9\ht\strutbox depth .9\dp\strutbox\relax}
\newcommand{\colorboxx}[2]{%
  \begingroup%
  \setlength{\fboxsep}{0pt}%
  \colorbox{#1}{\reducedstrut#2\/}%
  \endgroup
}
\newcommandx{\mgl}[2][1=]{\colorboxx{blue!30}{\tiny\,}\todo[color=blue!30,#1]{MG: #2}}
\newcommandx{\pb}[2][1=]{\colorboxx{gray!30}{\tiny\,}\todo[color=gray!30,#1]{PB: #2}}
\newcommandx{\cjd}[2][1=]{\colorboxx{cyan!30}{\tiny\,}\todo[color=cyan!30,#1]{CD: #2}}

\title{A Critical Analysis of Biased Parsers in Unsupervised Parsing}

\author{Chris Dyer \qquad G\'abor Melis \qquad Phil Blunsom \\
  DeepMind \\
  London, UK \\
  {\tt \{cdyer,melisgl,pblunsom\}@google.com}}

\date{}

\begin{document}
\maketitle
\begin{abstract}
A series of recent papers has used a parsing algorithm due to Shen et al.~(2018) to recover phrase-structure trees based on proxies for ``syntactic depth.'' These proxy depths are obtained from the representations learned by recurrent language models augmented with mechanisms that encourage the (unsupervised) discovery of hierarchical structure latent in natural language sentences. Using the same parser, we show that proxies derived from a conventional LSTM language model produce trees comparably well to the specialized architectures used in previous work. However, we also provide a detailed analysis of the parsing algorithm, showing (1)~that it is incomplete---that is, it can recover only a fraction of possible trees---and (2)~that it has a marked bias for right-branching structures which results in inflated performance in right-branching languages like English. Our analysis shows that evaluating with biased parsing algorithms can
inflate the apparent structural competence of language models.
\end{abstract}

\section{Introduction}

Several recent papers~\citep{shen:2018,htut:2018,shen:2019,li:2019,shi:2019} have used a new parsing algorithm to recover hierarchical constituency structures from sequiantial recurrent language models trained only on sequences of words. This parser, which we call a \pname parser for reasons which we will describe below, was introduced by \citet{shen:2018}. It operates top-down by recursively splitting larger constituents into smaller ones, based on estimated changes of ``syntactic depth'' until only terminal symbols remain~(\S\ref{sec:parser}). In contrast to previous work, which has mostly used explicit tree-structured models with stochastic latent variables, this line of work is an interestingly different approach to the classic problem of inferring hierarchical structure from flat sequences. 

In this paper, we make two primary contributions. First, we show, using the same methodology, that phrase structure can be recovered from conventional LSTMs comparably well as from the ``ordered neurons'' LSTM of \citet{shen:2019}, which is an LSTM variant designed to mimic certain aspects of phrasal syntactic processing~(\S\ref{sec:lstms}). Second, we provide a careful analysis of the parsing algorithm~(\S\ref{sec:analysis}), showing that in contrast to traditional parsers, which can recover any binary tree, the \pname parser is able to recover only a small fraction of all binary trees (and further, that many valid structures are not recoverable by this parser). This incomplete support, together with the greedy search procedure used in the algorithm, results in a marked bias for returning right-branching structures, which in turn overestimates parsing performance on right-branching languages like English.

Since an adequate model of syntax discovery must be able to account for languages with different branching biases and assign valid structural descriptions to any sentence, our analysis indicates particular care must be taken when relying on this algorithm to analyze the structural competence of language models.

\section{\pname Parsers}
\label{sec:parser}
Fig.~\ref{fig:parser} gives the greedy top-down parsing algorithm originally described in \citet{shen:2018} as a system of inference rules~\citep{goodman:1999}.\footnote{Our presentation is slightly different to the one in \citet{shen:2018}, but our notational variant was chosen to facilitate comparison with other parsers presented in these terms and also to simplify the proof of Prop. 2 (see \S\ref{sec:analysis}).} We call this a \pname parser, which will be explained below in the analysis section~(\S\ref{sec:analysis}). The parser operates on a sentence of length $n$ by taking a vector of scores $\mathbf{s} \in \mathbb{R}^n$ and recursively splitting the sentence into pairs of smaller and smaller constituents until only single terminals remain. Given a constituent $[i,j]$ and the score vector, only a single inference rule ever applies. When the deduction completes, the collection of constituents encountered during parsing constitutes the parse.

\begin{figure*}
\begin{align*}
&    \textbf{Premises} & \frac{}{[1,n]} \\
&    \textbf{Inference rules} \\
&\textsc{binary} &  \frac{[i,j]}{[i,i]\ [j,j]} & \quad j-i=1 \\
&\textsc{left} & \frac{[i,j]}{[i,i] \ [i+1, j]} & \quad j-i > 1 \wedge i=\arg \max_{\ell \in [i,j]} s_{\ell} \\
&\textsc{right} & \frac{[i,j]}{[i,j-1] \ [j, j]} & \quad j-i > 1 \wedge j=\arg \max_{\ell \in [i,j]} s_{\ell} \\
&\textsc{middle} & \frac{[i,j]}{[i,k-1] \ [k, j]} & \quad j-i > 1 \wedge k=\arg \max_{\ell \in [i,j]} s_{\ell} \wedge k \in [i+1,j-1] \\
& \textbf{Goals} & [i,i] & \quad \forall i \in [1,n]
\end{align*}
\vspace{-.9cm}\caption{The \pname parser as a system of inference rules. Inputs are a vector of scores $\mathbf{s} \in \mathbb{R}^n$ where $s_i$ is the score of the $i$th word. An item $[i,j]$ indicates a consistuent spans words $i$ to $j$, inclusive. Inference rules are applied as follows: when a constituent matching the item above the line, subject to the constraints on the right is found, then the constituents below the line are constructed. The process repeats until all goals are constructed.
\label{fig:parser}}
\end{figure*}

Depending on the interpretation of $s_i$, the consequent of the \textsc{middle} rule can be changed to be $[i,k]\ [k+1, j]$, which places the word triggering the split of $[i,j]$ in the left (rather than right) child constituent. We refer to these variants as the L- and R-variants of the parser. We analyze the algorithm, and the implications of the these variants below~(\S\ref{sec:analysis}), but first we provide a demonstration of how this algorithm can be used to recover trees from sequential neural language models.

\section{Unsupervised Syntax in LSTMs}
\label{sec:lstms}
\citet{shen:2019} propose a modification to LSTMs that imposes an ordering on the memory cells. Rather than computing each forget gate independently given the previous state and input as is done in conventional LSTMs, the forget gates in Shen et al.'s ordered neuron LSTM (ON-LSTM) are tied via a new activation function called a cumulative softmax so that when a higher level forget gate is on, lower level ones are forced to be on as well. The value for $s_i$ is then defined to be the average forget depth---that is, the average number of neurons that are turned off by the cumulative softmax---at each position $i \in [1,n]$. Intuitively, this means that closing more constituents in the tree is linked to forgetting more information about the history. In this experiment, we operationalize the same linking hypothesis, but we simply sum the values of the forget gates at each time step (variously at different layers, or all layers together) in a conventional LSTM to obtain a measure of how much information is being forgotten when updating the recurrent representation with the most recently generated word. To ensure a fair comparison, both models were trained on the 10k vocabulary Penn Treebank (PTB) to minimize cross entropy; they made use of the same number of parameters; and the best model was selected based on validation perplexity. Details of our LSTM language model are found in Appendix~\ref{sec:lstm_details}.

Results on the 7422 sentence WSJ10 set (consisting of sentences from the PTB of up to 10 words) and the PTB23 set (consisting of all sentences from \S23, the traditional PTB test set for supervised parse evaluation) are shown in Tab.~\ref{tab:f1}, with the ON-LSTM of \citet{shen:2019} provided as a reference. We note three things. First, the F1 scores of the two models are comparable; however, the LSTM baseline is slightly worse on WSJ10. Second, whereas the ON-LSTM model requires that a single layer be selected to compute the parser's scoring criterion (since each layer will have a different expected forget depth), the unordered nature of LSTM forget gates means our baseline can use gates from all layers jointly. Third, the L-variant of the parser is substantially worse.

\begin{table}[t]
    \centering
    \vspace{-4mm}
    \begin{tabular}{l c r r}
    
    \toprule
    & &  WSJ10 &  PTB23 \\
    \midrule
ON-LSTM-1$^{\dagger}$ & R & 35.2 & 20.0 \\
ON-LSTM-2$^{\dagger}$ & R & \textbf{65.1} & \textbf{47.7} \\
ON-LSTM-3$^{\dagger}$ & R & 54.0 & 36.6 \\
\midrule
LSTM-1 & R & 58.4 & 43.7 \\
LSTM-2 & R & 58.4 & 45.1 \\
LSTM-1,2 & R & \textbf{60.1} & \textbf{47.0} \\
\midrule
LSTM-1 & L & 43.8 & 31.8 \\
LSTM-2 & L & 47.4 & 35.1 \\
LSTM-1,2 & L & 46.3 & 33.8 \\
 \bottomrule
    \end{tabular}
    \caption{$F_1$ scores using the same evaluation setup as \citet{shen:2019}. Numbers in the model name give the layer $\mathbf{s}$ was extracted from. R/L indicates which variant of the parsing algorithm was used. Results with $^{\dagger}$ are reproduced from Table 2 of \citet{shen:2019}.}
    \label{tab:f1}
\end{table}

\section{Analysis of the \pname Parser}
\label{sec:analysis}
We now turn to a formal analysis of the parser, and we show two things: first, that the parser is able to recover only a rapidly decreasing (with length) fraction of valid trees~(\S\ref{sec:coverage}); and second, that the parser has a marked bias in favour of returning right-branching structures~(\S\ref{sec:bias}).

\subsection{Incomplete support}
\label{sec:coverage}
Here we characterize properties of the trees that are recoverable with the \pname parser, and describe our reason for naming it as we have.
\begin{prop} Ignoring all single-terminal bracketings, the R-variant \pname parser can generate all valid binary bracketings that do not contain the contiguous string $)(($.\footnote{Proofs of propositions are found in Appendix~\ref{sec:proofs}.}
\end{prop}

\begin{figure}[h]
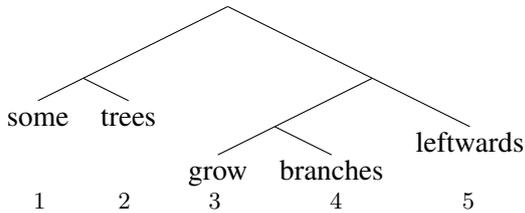

\Tree [ [ some trees ] [ [ grow branches ] leftwards ] ]

\noindent \begin{footnotesize}
\hspace{-0.6cm}$\ \quad 1 \qquad \ \ \ \     2 \qquad \ \ \ \ \  3 \qquad \qquad \ \; 4 \qquad \qquad \quad 5$
\end{footnotesize}
    \centering
    \vspace{-.15cm}\caption{Example of an unrecoverable tree structure and possible sentence with that structure, \emph{((some trees) ((grow branches) leftwards)}, which includes the forbidden string $)(($.}
    \label{fig:impossible_tree}
\end{figure}

This avoidance of \textbf{c}lose-\textbf{o}pen-\textbf{o}pen leads us to call this the \pname parser, where the notation $\overline{x}$ indicates that $x$ is forbidden. In Fig.~\ref{fig:impossible_tree} we give an example of an unrecoverable parse for a sentence of $n=5$ (there are 14 binary bracketings of length-5 sentences, but the \pname parser can only recover 13 of them).

\begin{prop} The number of parses of a string of length $n$ that is recoverable by a $\coo$ parser is given by $a_{n}$\footnote{This sequence is \url{https://oeis.org/A082582}, which counts a permutation-avoiding path construction that shows up in several combinatorial problems~\citep{baxter:2015}.}, where
\begin{align*}
    a_1 &= 1 \qquad  \qquad a_2 = 1 \\
    a_n &= 2a_{n-1} + \sum_{k=2}^{n-1} a_{k-1} \times a_{n-k}.
\end{align*}
\end{prop}

\begin{figure}
    \centering
    \includegraphics[width=0.48\textwidth]{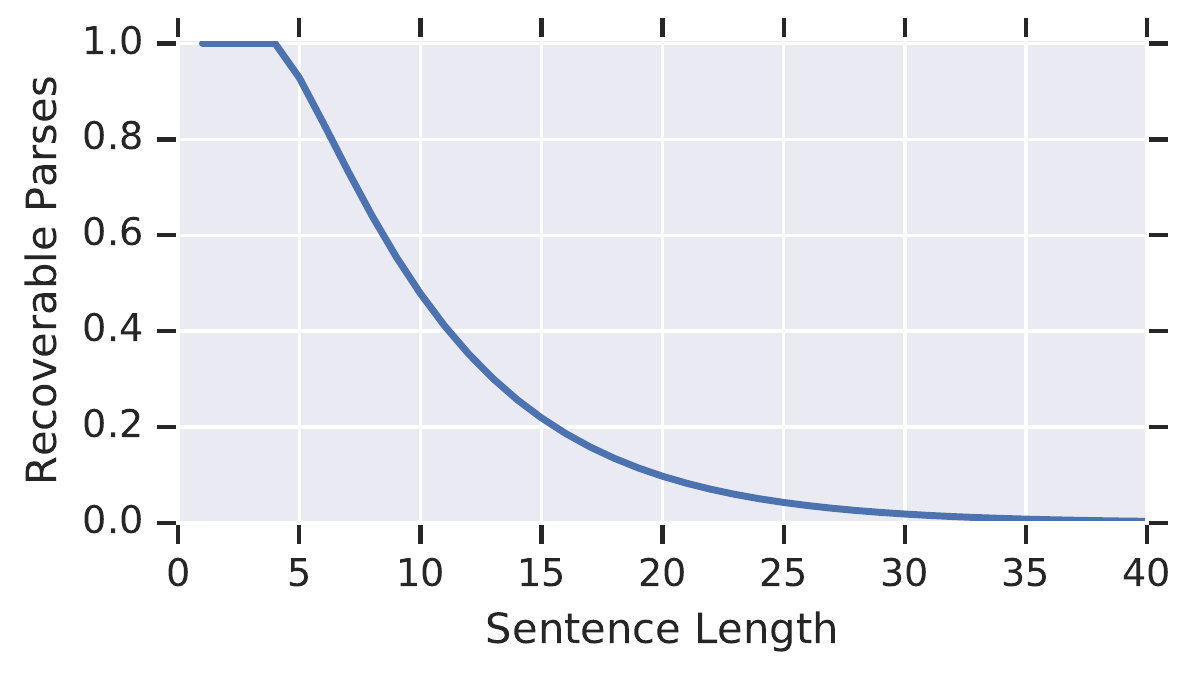}
    \vspace{-.75cm}\caption{The proportion of valid parses recoverable as a function of sentence length, i.e., $a_n/C_{n-1}$.}
    \label{fig:ratio}
\end{figure}

Although this sequence grows in $\Theta(2^n)$, Fig.~\ref{fig:ratio} shows that as the length $n$ of the input increases, the ratio of the extractable parses to the number of total binary parses, which is given by the $(n-1)$th Catalan number, $C_{n-1}$ \cite{motzkin:1948}, converges logarithmically to 0.

\subsection{Branching direction bias}
\label{sec:bias}
\begin{figure}
    \centering
    \includegraphics[width=0.48\textwidth,trim=0 180 0 0,clip]{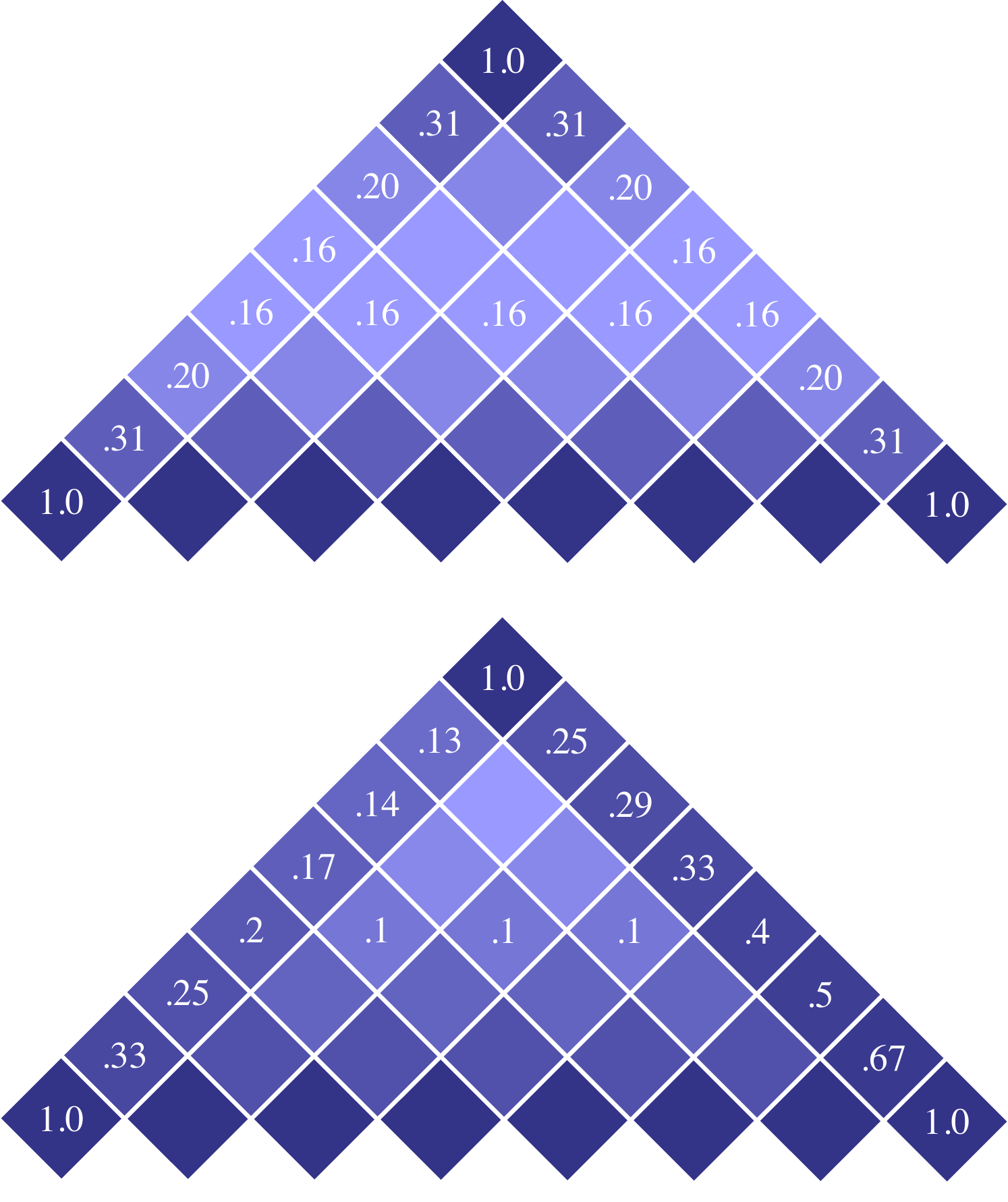}
    \vspace{-.75cm}\caption{Probability that $[i,j]$ is a constituent when all binary trees are eqiprobable~(above), and when all relative orders of syntactic depths are equiprobable, when decoded by the R-\pname parser~(below).}
    \label{fig:prior}
\end{figure}

Since not every binary tree can be recovered by a \pname parser, we explore to what extent this biases the solutions found by the algorithm.
We explore the nature of the bias in two ways. First, in Fig.~\ref{fig:prior} we plot the marginal probability that $[i,j]$ is a constituent when all binary trees are equiprobable, and compare it with the probability that the R-\pname parser will identify it as a span under a uniform distribution over the relative orderings of the $s_i$'s (since it is the relative orderings that determine the parse structure).
As we can see, there is no directionality bias in the uniform tree model (all rows have constant probabilities), but in the R-\pname parser, the probability of the right-most constituent $[n-\ell+1, n]$ of length $1 < \ell < n$ is twice that of the left-most one $[1, \ell]$, indicating a right-branching bias. The L-variant marginals (not shown) are the reflection of the R-variant ones across the vertical axis, indicating it has the same bias, only reversed in direction.

Thus, while the R-variant \pname parser fails to reach a large number of trees, it nevertheless has a bias toward the right-branching syntactic structures that are common in English. Since parse evaluation is in terms of retrieval of spans (not entire trees), we may reasonably assume that the right-branching bias is more beneficial than the existence of unreachable trees (correct though they may be) is harmful. A final experiment supports this hypothesis: we run a Monte Carlo simulation to approximate the expected F1 score obtained on WSJ10 and PTB23 under the uniform binary distribution, the left-skewed distribution, and the right-skewed distribution. These estimates are reported in Tab.~\ref{tab:mc} and show that the R-variant parser confers significant advantages over the uniform tree model, and the L-variant parser is worse again.

\begin{table}[h]
    \centering
    \begin{tabular}{l c c}
    
    \toprule
    &  WSJ10 &  PTB23 \\
    \midrule
Left-skew & 31.6 (${\footnotesize \sigma=0.2}$) & 16.8 (${\footnotesize \sigma=0.1}$) \\
Uniform  & 33.7 (${\footnotesize \sigma=0.2}$) & 18.3 (${\footnotesize \sigma=0.1}$) \\
Right-skew & \textbf{37.5} (${\footnotesize \sigma=0.2}$) & \textbf{19.9} (${\footnotesize \sigma=0.2}$) \\
 \bottomrule
 \end{tabular}
  \caption{Expected F1 score under different distributions of trees. \label{tab:mc}}
\end{table}

\section{Potential Fixes}
Is it possible to fix the parser to remove the biases identified above and to make the parser complete? It is. In general, it is possible to recursively split phrases top down and to obtain any possible binary bracketing~\cite{stern:2017,shen:acl:2018}. The locus of the bias in the $\coo$ parser is the decision rule for splitting a span into two daughters, based on the maximally ``deep'' word. Since the maximum word in a larger span will still be maximal in a resulting smaller span, certain configurations will necessarily be unreachable. However, at least two alternative scoring possibilities suggest themselves: (1)~scoring transitions between words rather than individual words and (2)~scoring spans rather than words. By scoring the $O(n)$ transitions between words, each potential division between $i,i+1$ will lie outside of the resulting daughter spans, avoiding the problem of having the maximally scoring element present in both the parent and the daughter. By scoring the $O(n^2)$ spans rather than words or word gaps, a similar argument holds.

Although these algorithms are well known, they have only been used for supervised parsing. The challenge for using either of these decision rules in the context of unsupervised parsing is to find a suitable linking hypothesis that connects the activations of a sequential language model to hypotheses about either changes in syntactic depth that occur from one word to the next (i.e., scoring the gaps) or that assign scores to all spans in a sequence. Candidates for such quantities in conventional networks do not, however, immediately suggest themselves.

\section{Related Work}
Searching for latent linguistic structure in the activations of neural networks that were trained on surface strings has a long history, although the correlates of syntactic structure have only recently begun to be explored in earnest. \citet{elman:1990} used entropy spikes to segment character sequences into words and, quite similarly to this work, \citet{wang:2017} used changes in reset gates in GRU-based autoencoders of acoustic signals to discover phone boundaries. \citet{hewitt:2019} found they could use linear models to recover the (squared) tree distance between pairs of words as well as depth in the contextual embeddings of undirected language models. 

A large number of papers have also used recursively structured networks to discover syntax, refer to a \citet{shen:2019} for a comprehensive list.

\section{Discussion}

The learning process that permits children to acquire a robust knowledge of hierarchical structure from unstructured strings of words in their environments has been an area of active study for over half a century. Despite the scientific effort---not to mention the reliable success that children learning their first language exhibit---we still have no adequate model of this process. Thus, new modeling approaches, like the ones reviewed in this paper, are important. However, an understanding of what the results are telling us is more than simply a question of F1 scores. In this case, a biased parser that is well-matched to English structures appears to show that LSTMs operate more syntactically than they may actually do. Of course, the tendency for languages to have consistent branching patterns has been hypothesized to arise from a head-directionality parameter that is set during learning~\citep{chomsky:1981}, and biases of the sort imposed by the \pname parser could be reasonable, if a mechanism for learning the head-directionality parameter were specified.\footnote{Indeed, children demonstrate knowledge of word order from their earliest utterances~\citep{brown:1973}, and even pre-lexical infants are aware of word order patterns in their first language before knowing its words~\citep{gervain:2008}.} 
When comparing unsupervised parsing algorithms aiming to shed light on how children discover structure in language, we must acknowledge that the goal is not simply to obtain the best accuracy, but to do so in a plausibly language agnostic way. 
If one's goal is not to understand the general problem of structure, but merely to get better parsers, language specific biases are certainly on the table. However, we argue that the work should make these goals clear, and that it is unreasonable to mix comparisons across models designed with different goals in mind.

\ignore{
\paragraph{Can the parser be fixed?} \mgl{The parser need not necessarily be fixed if the goal is to improve parsing (of English). I thought the problem was that they used the biased parser's results to make claims about model bias.}
\pb{Given the fact that the parser is deterministic and can't find all paths, I think it is fair to characterise it as broken. The bias is a separate question of what hypothesis is being tested. Klein and Manning had one, but Shen et al. do not.}
There are similar algorithms for top-down parsing that do not impose the same biases as the \pname parser. \citet{stern:2017} present an algorithm that likewise recursively splits spans $[i,j]$; however, crucially the splitting figure of merit is not merely a function of the sentence and the position; rather it depends on $i$ and $j$, meaning that while a higher level sentence may split $k$; the resulting subspan $[k,j]$ can freely chose where to split again (including, of course, splitting $k$ off as a terminal, as is obligatory in the \pname parser). Alternatively, by assigning scores to the ``spaces'' between word, and splitting on those, there is no preference for either left- or right-branching structures; however, the nature of a suitable linking hypothesis between the dynamics of recurrent language models and syntactic depths of transitions from one state to the next is uncertain.
}

This paper has drawn attention to a potential confound in interpreting the results of experiments using \pname parsers. We wish to emphasize again that the works employing this parser represent a meaningful step forward in the important scientific question of how hierarchical structure is inferred from unannotated sentences. However, an awareness of the biases of the algorithms being used to assess this question is important.

\section*{Acknowledgments}
We thank Adhi Kuncoro for his helpful suggestions on this writeup.

\bibliography{main}
\bibliographystyle{acl_natbib}

\newpage

\appendix
\ignore{
\section{Context-Free Grammar Equivalent}
\label{sec:grammar}
We provide a characterization of the constraints on the \pname parsing algorithm in terms of a context-free grammar. Parsing with this grammar yields the same set of brackets of a string.
\begin{align*}
    \textrm{S} &\rightarrow \textrm{T} \\
    \textrm{S} &\rightarrow \textrm{T S} \\
    \textrm{S} &\rightarrow \textrm{S T} \\
    \textrm{S} &\rightarrow \textrm{S Z} \\
    \textrm{Z} &\rightarrow \textrm{T S} \\
    \textrm{T} &\rightarrow w & \forall w \in \Sigma \\
\end{align*}
By converting this to Chomsky normal form, this grammar can be used to impose the same structural constraints on other binary bracketing grammars by jointly parsing with both.
}
\section{Experimental Details}
\label{sec:lstm_details}
For our experiments, we trained a two-layer LSTM with 950 units and 24M parameters on the Penn Treebank~\citep[][PTB]{marcus1993building} language modelling dataset with preprocessing from \citet{mikolov2010recurrent}. Its lack of architectural innovations makes this model ideal as a baseline. In particular, we refrain from using recent inventions such as the Mixture of Softmaxes \citep{yang2017breaking} and FRAGE \citep{gong2018frage}. Still, the trained model is within 2 perplexity points of the state of the art \citep{gong2018frage}, achieving 55.8 and 54.6 on the validation and test sets, respectively.

As is the case for all models that are strong on small datasets such as the PTB, this one is also heavily regularized. An $\ell_2$ penalty controls the magnitude of all weights. Dropout is applied to the inputs of the two LSTM layers, to their recurrent connections \citep{gal2016theoretically}, and to the output. Optimization is performed by Adam \citep{kingma2014adam} with $\beta_1=0$, a setting that resembles RMSProp without momentum. Dropout is turned off at test time, when we extract values of the forget gate.

\section{Proofs of Propositions}
\label{sec:proofs}
Although it was convenient to present the results in reverse order, we first prove Proposition 2, the recurrence counting the number of recoverable parses, and then Proposition 1 since it is used in proving Proposition 1.
\subsection{Proof of Proposition 2}
To derive the recurrence giving the number of recoverable parses, the general strategy will be to exploit the duality between top-down and bottom-up parsing algorithms by reversing the inference system in Fig.~\ref{fig:parser} and parsing sentences bottom up. This admits a more analyzable algorithm. In this proof, we focus on the default R-variant, although the analysis is identical for the L-variant.

We begin by noting that after the top-down \textsc{middle} rule applies, the new constituent $[k,j]$ will be used immediately to derive a pair of constituents $[k,k]$ and $[k+1,j]$ via either the \textsc{left} or \textsc{binary} rule, depending on the size of $[k,j]$. This is because $k$ is the index of the maximum score in $[i,j]$, therefore it will also be the index of the maximum in the daughter constituent $[k,j]$. Since there is only a single derivation of $[k,j]$ from its subconstituents (and vice-versa), we can combine these two steps into a single step by transforming the \textsc{middle} rule as follows:
\begin{align*}
&\textsc{middle}' &   \frac{[i,j]}{[i,k-1] \ [k,k] \ [k+1, j]}
\end{align*}
\begin{align*}
    \textrm{when } j-i > 1 &\wedge k=\arg \max_{l \in [i,j]} s_l \\
                           &\wedge k \in [i+1,j-1].
\end{align*}
This ternary form results in a system with the same number of derivations but is easier to analyze.

In the bottom-up version of the parser, we start with the goals of the top-down parser, run the inference rules backwards (from bottom to top) and derive ultimately the top-down parser's premise as the unique goal. Since we want to consider all possible decisions the parser might make, the bottom-up rules are also transformed to disregard the constraints imposed by the weight vector. Finally, to count the derivations, we use an inside algorithm with each item associated with a number that counts the unique derivations of that item in the bottom-up forest. When combining items to build a new item, the weights of the antecedent items multiply; when an item may be derived multiple ways, they add~\citep{goodman:1999}.

Let $a_n$ refer to the number of possible parses for a sequence of length $n$. It is obvious that $a_1=1$ since the only way to construct single-length items in the bottom-up parser (i.e., $[i,i]$) is with the initialization step (all other rules have constraints that build longer constituents). Likewise $a_2=1$ since there is only one way to build a constituent of length 2, namely the \textsc{binary} rule, which combines two single-length constituents (each which can only be derived one way).

Consider the general case $a_n$ where $n>2$. Here, there are three ways of deriving $[1,n]$: the \textsc{left} and \textsc{right} rules, and the more general $\textsc{middle}'$ rule. Both \textsc{left} and \textsc{right} combine a tree spanning $n-1$ symbols with a single terminal. The single terminal has, as we have seen, one derivation, and the tree has, by induction, $a_{n-1}$ derivations. Thus, the \textsc{left} and \textsc{right} rules contribute $2a_{n-1}$ derivations to $a_n$.

It remains to account for the contribution of the $\textsc{middle}'$ rule. To do so, we observe that $\textsc{middle}'$ derives a span $[1,n]$ by combining two arbitrarily large spans: $[1,k-1]$ and $[k+1,n]$, having derivation counts $a_{k-1}$ and $a_{n-k}$ respectively (by induction), with a single element $[k,k]$, having a single derivation (by definition). Thus, for a possible value of $k$, the contribution to $a_n$ is $a_{k-1} \times a_{n-k}$. Since $k$ may be any value in the range $[2,n-1]$, we have $\sum_{k=2}^{n-1} a_{k-1} \times a_{n-k-1}$ in aggregate as the contribution of $\textsc{middle}'$. Thus, combining the contributions of \textsc{left}, \textsc{right}, and $\textsc{middle}'$, we obtain:
\begin{align*}
    a_n = 2a_{n-1} + \sum_{k=2}^{n-1} a_{k-1} \times a_{n-k},
\end{align*}
for $n > 2$.\qed

\subsection{Proof of Proposition 1}
We prove this in two parts, first we show that the string $)(($ cannot be generated by the parser. Second we show that when brackets containing this string are removed from the set of all binary brackets of $n$ symbols, its cardinality is $a_n$.

\paragraph{Part 1.} We prove that the string $)(($ cannot be generated by the parser by contradiction. Assume that the bracketing produced by the parser contains the string $)(($. To exist in a balanced binary tree, there must be at least two symbols to the left (terminals are unbracketed, therefore the closing bracket that has not ended the tree will be at least a length-2 constituent); and three symbols to the right of the split (if the following material was length-2, then two opening brackets would result in a unary production, which cannot exist in a binary tree).

Thus, to obtain the split between $)$ and $($, the \textsc{middle} rule would have applied because of the size constraints on the inference rules. However, as we showed in the proof of Prop.\ 1, after the \textsc{middle} applies, a terminal symbol must be the left child of the resulting right daughter constituent, thus we obligatorily must have the sequence $)(x($, where $x$ is any single terminal symbol. This contradicts our starting assumption. \qed

\paragraph{Part 2.} It remains to show that the only unreachable trees are those that contain $)(($. Proving this formally is considerably more involved than Part 1, so we only outline the strategy. First, we construct a finite state machine that generates a language in $\{ (, ), x \}$ consisting of all strings that do not contain the contiguous substring $)(($. Second, we intersect this with the following grammar that generates all valid binary bracketed expressions:
\begin{align*}
    S &\rightarrow (\textrm{S S}) \\
    S &\rightarrow x
\end{align*}
Finally, we remove the bracketing symbols, and show that the number of derivations of the string $x^n$ is $a_n$.
\qed

\end{document}